\documentclass[submission,copyright,creativecommons]{eptcs}
\providecommand{\event}{ICLP 2021} 

\usepackage{mathptmx}
\usepackage{breakurl}             
\usepackage{underscore}           
\usepackage{hyperref}
\usepackage{mathtools}
\usepackage{amssymb}
\usepackage{bm}
\usepackage{amsmath}
\usepackage{xspace}
\usepackage{latexsym}

\newcommand{\bnf}  { \Coloneqq }
\renewcommand{\phi}  { \varphi }
\newcommand{\lang}[1]  { {\cal L}_{#1} }

\newcommand{\card}[1]{\left\vert{#1}\right\vert}
\newcommand{\tuple}[1]  { \langle #1 \rangle }
\newcommand{\classnot}{{\sim}}
\newcommand{\Prop}  { \mathbb{P} }

\newcommand{\negof}[1]{\widetilde{#1}}
\newcommand{\suchthat}  { \ :\ }
\newcommand\restr[2]  { \ensuremath{\left.#1\right|_{#2}} }

\newcommand{\quotes}[1]  {``#1''}
\newcommand{\singlequote}[1]  {`#1'}
\newcommand{\doublequotes}[1]  {``#1''}
\newcommand{\onequote}[1]  {`#1'}
\newcommand{\set}[1]  { \{ #1 \} }
\newcommand{\Set}[1]  { \big\{ #1 \big\} }

\newcommand{\K}{ \mathsf{K} }
\newcommand{\M}{ \mathsf{M} }

\newcommand{\ESmodels}{\models_{\scriptscriptstyle{\ES}}}
\newcommand{\notESmodels}{\not\models_{\scriptscriptstyle{\ES}}}

\newcommand{\epispec}{\Pi}
\newcommand{\anset}{\mathtt{AS}}
\newcommand{\epset}{\texttt{Ep}}
\newcommand{\lpnot}{ \mathtt{not} \,}
\newcommand{\notlp}{ \mathtt{not} }

\newcommand{\Notlp}{ \mathtt{NOT} }
\newcommand{\lpor}{ \, \mathtt{or} \,}
\newcommand{\orlp}{ \mathtt{or} }

\newcommand{\weak}{\mathtt{s}}
\newcommand{\weakrel}{\mathtt{s_{r}}}

\newcommand{\intimp}{\rightarrow}

\newcommand{\intnot}{\neg}

\newcommand{\sfivemodels}{\models_{\scriptscriptstyle{\sfive}}}

\newcommand{\eemset}{\mathtt{EEM}}

\newcommand{\eemsetonbes}{\mathtt{EEM_{\scriptscriptstyle{15}}}\!}
\newcommand{\aeemsetonbes}{\mathtt{AEEM_{\scriptscriptstyle{15}}}\!}
\newcommand{\eemsetyirmi}{\mathtt{EEM_{\scriptscriptstyle{20}}}\!}
\newcommand{\aeemsetyirmi}{\mathtt{AEEM_{\scriptscriptstyle{20}}}\!}
\newcommand{\eemsetyirmibir}{\mathtt{EEM_{\scriptscriptstyle{21}}}\!}
\newcommand{\aeemsetyirmibir}{\mathtt{AEEM_{\scriptscriptstyle{21}}}\!}

\newcommand{\Ehtmodels}{ \models_{\scriptscriptstyle \Eht} }
\newcommand{\notEhtmodels}{ \not\models_{\scriptscriptstyle \Eht} }

\newcommand{\Ehtmodelsyirmi}{ \models_{\scriptscriptstyle{\EHTyirmi}} }

\newcommand{\Ehtmodelsyirmibir}{ \models_{\scriptscriptstyle{\EHTyirmibir}} }
\newcommand{\notEhtmodelsyirmibir}{ \not\models_{\scriptscriptstyle{\EHTyirmibir}} }

\newcommand{\there}{\mathcal{A}}
\newcommand{\thereb}{\mathcal{A'}}

\newcommand{\starmodels}{ \models^*}

\newcommand{\trOf}[1]{#1 ^*}
\newcommand{\Khat}{ \hat{\mathsf{K}}}

\newcommand{\Eht}{\ensuremath{\mathsf{EHT}}\xspace}

\newcommand{\sfive}{\ensuremath{\mathsf{S5}}\xspace}

\newcommand{\Ht}{\ensuremath{\mathsf{HT}}\xspace}

\newcommand{\ASP}{\ensuremath{\mathsf{ASP}}\xspace}
\newcommand{\EL}{\ensuremath{\mathsf{EL}}\xspace}

\newcommand{\ES}{\ensuremath{\mathsf{ES}}\xspace}

\newcommand{\ESdoksanbir}  { \logic{ES_{\scriptscriptstyle{91}} }}
\newcommand{\ESdoksandort}  { \logic{ES_{\scriptscriptstyle{94}} }}
\newcommand{\ESonbir}  { \logic{ES_{\scriptscriptstyle{11}} }}
\newcommand{\ESondort}  { \logic{ES_{\scriptscriptstyle{14}} }}
\newcommand{\ESonbes} { \logic{ES_{\scriptscriptstyle{15}} }}
\newcommand{\ESonalti}  { \logic{ES_{\scriptscriptstyle{16}} }}
\newcommand{\ESonsekiz}  { \logic{ES_{\scriptscriptstyle{18}} }}

\newcommand{\ESyirmi}  { \logic{ES_{\scriptscriptstyle{20}} }}
\newcommand{\ESyirmibir}  { \logic{ES_{\scriptscriptstyle{21}} }}
\newcommand{\ESx}  { \logic{ES_{\scriptscriptstyle{X}} }}
\newcommand{\EHTonbes}  { \logic{EHT_{\!\!\scriptscriptstyle{15}} }}
\newcommand{\EHTyirmi}  { \logic{EHT_{\!\!\scriptscriptstyle{20}} }}
\newcommand{\EHTyirmibir}  { \logic{EHT_{\!\!\scriptscriptstyle{21}} }}
\newcommand{\EHTonbesmodel}  { \mathtt{EHT_{\!\scriptscriptstyle{15}} }\!}
\newcommand{\EHTyirmimodel}  { \mathtt{EHT_{\!\scriptscriptstyle{20}} }\!}
\newcommand{\EHTyirmibirmodel}  { \mathtt{EHT_{\!\scriptscriptstyle{21}} }\!}

\newcommand{\EEMyirmi}  { \mathtt{EEM_{\scriptscriptstyle{20}} }\!}

\newcommand{\AEEMyirmi}  { \mathtt{AEEM_{\scriptscriptstyle{20}} }\!}
\newcommand{\AEEMyirmibir}  { \mathtt{AEEM_{\scriptscriptstyle{21}} }\!}
\newcommand{\logic}[1]  { \ensuremath{\mathsf{#1}}  \xspace }
\newcommand{\KD}  { \logic{KD45} }

\newcommand{\SW}  { \logic{SW5} }

\newcommand{\ael}  { \logic{AEL} }
\newcommand{\rael}  { \logic{RAEL} }

\newtheorem{thm}{Theorem}

\newtheorem{prop}{Proposition}

\newtheorem{coroll}{Corollary}

\newtheorem{example}[thm]{Example}
\newcommand{\eqfp}{\stackrel{\mathsf{fp}}{=}}
\newcommand{\eqdef}{\stackrel{\mathsf{def}}{=}}
\newcommand{\naf}{\texttt{NAF}}

\title{Refining the Semantics of Epistemic Specifications}
\author{Ezgi Iraz Su\thanks{I sincerely thank the anonymous reviewers for taking the time and effort to give some useful
comments and suggestions about the earlier draft. I also wish to thank the program chairs for their help and understanding
in submitting the final version.}
\institute{Sinop University, Department of Computer Engineering, Sinop, Turkey}
\email{eirazsu@sinop.edu.tr}
}

\def\titlerunning{Refining the semantics of \texorpdfstring{$\ES$}{ES}}
\def\authorrunning{E.I. Su}

\begin{document}
\maketitle

\begin{abstract}
\emph{Answer set programming} ($\ASP$) is a problem-solving approach, 
which has been strongly supported both scientifically
and technologically by several solvers, ongoing active research, and implementations
in many different fields.
However, although researchers acknowledged long ago the necessity of epistemic
operators in the language of $\ASP$ 
for better introspective reasoning, this research venue did not attract much attention
until recently.
Moreover, the existing epistemic
extensions of $\ASP$ in the literature are not widely approved either, due to the fact
that some propose unintended results even for some simple
acyclic epistemic programs, new unexpected results may
possibly be found, and more importantly, researchers have
different reasonings for some critical programs.
To that end, Cabalar et al.\ have recently identified 
some structural properties of epistemic programs to 
formally support a possible semantics proposal of such programs and standardise their results.
Nonetheless,
the soundness of these properties is still under debate, and
they are not widely accepted either by the $\ASP$ community. 
Thus, it seems that there is still time to really understand the paradigm, have a mature formalism, 
and determine the principles providing formal justification of their understandable models.
In this paper, we mainly focus on the existing semantics approaches, 
the criteria that a satisfactory semantics is supposed to satisfy, and
the ways to improve them.
We also extend some well-known propositions of here-and-there logic ($\Ht$) into epistemic $\Ht$ so as to reveal the
real behaviour of programs.
Finally, we propose a slightly novel semantics for epistemic $\ASP$, which can be considered as a reflexive extension of
Cabalar et al.'s recent formalism called autoepistemic $\ASP$.
\end{abstract}


\section{Introduction}  \label{sec:Introduction}
\emph{Answer set programming} ($\ASP$) has been proposed by Gelfond and Lifschitz (GL)
\cite{GelfondL88} as an approach to declarative programming. Its reduct-based GL-semantics
is given by
\emph{answer sets} (alias, \emph{stable models})---consistent sets $A$ of ground literals\footnote{The use of variables
in $\ASP$-programs
is understood as abbreviations for the collection of their ground (variable-free) instances. Thus, for simplicity,
in this paper we restrict the language of (epistemic) $\ASP$ to the propositional case. In $\ASP$,
a ground literal is a propositional variable (here, referred to as an \emph{atom}) $p$ or a
\emph{strongly-negated} propositional variable $\classnot p$.} (referred to as
\emph{valuations})
in which $p \notin A$ or $\classnot p \notin A$ for every atom $p$,
roughly described as the smallest per subset relation, and supported classical models of a program.
$\ASP$ provides a successful, and relatively simple way of solving
problems: first, a problem is encoded as a logic program whose answer sets correspond to solutions. Then,
by means of efficient $\ASP$-solvers computing these models, we obtain solutions in the form of answer sets.
As a result, currently, $\ASP$ has a wide range of applications in science and technology.
However, as first recognised by Gelfond \cite{Gelfond91strong},
$\ASP$ is not strong enough to correctly reason about the global situation
in the presence of multiple answer sets of a program
and then to derive new results out of the incomplete information these answer
sets convey altogether. One reason for this drawback is
the local performance of the $\ASP$'s negation as failure ($\texttt{NAF}$) operator (aka, default negation):
note that $\naf$ can only reflect incomplete information of each answer set individually,
but in order to extend the issue to the whole range of answer sets for global reasoning, we need epistemic
modal operators, which are able to quantify over a collection of answer sets.

The first approach of this line of research is Gelfond's \emph{epistemic specifications}
($\ESdoksanbir$) \cite{Gelfond91strong,Gelfond94}: he extended $\ASP$
with epistemic constructs called \emph{subjective literals}.
Indeed, with the inclusion of the epistemic modalities $\K$ and $\M$ (respectively having the literal readings
\quotes{\emph{known}}
and \quotes{\emph{may be believed}} in $\ESdoksanbir$), he could encode information of answer set collections.
The interpretation of this new language was in terms of
\emph{world-views}---collections $\mathcal A$ of valuations $A$, each of which constitutes
a minimal pointed classical S5-model\footnote{Particularly here, we regard S5-models 
as cluster structures in which every world is related to any other, including itself.} 
$(\mathcal A, A)$ of a program $\epispec$ w.r.t.\ truth and knowledge.
%
Similarly to answer sets, world-views are also reduct-based.
The reduct definition of the former eliminates default-negated constructs (i.e., $\texttt{NAF}$) w.r.t.\ a candidate answer set $A$ so that
the reduct is a positive $\ASP$-program, excluding $\naf$;
whereas the goal of the latter in $\ESdoksanbir$ is, in principle, to remove epistemic
constructs w.r.t.\ a candidate world-view $\mathcal A$. Thus, the
resulting program $\epispec^{\mathcal A}$ appears to be a regular $\ASP$-program, possibly including $\naf$
(but excluding $\K$ and $\M$). 
Then, we generate the collection $\mathcal A'$ of
all answer sets of this reduct $\epispec^{\mathcal A}$. Finally, if $\mathcal A'$ equals our candidate model $\mathcal A$, then
we call $\mathcal A$ a \emph{world-view} of the original program $\epispec$.

Researchers have soon realised that $\ESdoksanbir$ allows unsupported world-views.
Then, not only Gelfond himself \cite{gelfond2011new}, but also many others 
have come up with several different semantics proposals for epistemic
specifications ($\ES$); one following the other in order to get rid of newly-appearing unintended results.
The majority \cite{kahl2014refining,
ShenEiter16,KahlLS16,Kahlwvconstraints18,Su19jelia,Su19revisiting} 
are reduct-based world-view semantics.
The rest \cite{WangZ05nested,
Su15,SuFH20,CabalarAEEL20}
are inspired by  
Pearce's equilibrium-model
approach \cite{Pearce06}, characterising answer sets on a purely logical domain through 
minimal model reasoning.
They are based on epistemic extensions of equilibrium logic.

Up to recently, novel formalisms of $\ES$ were basically tested in terms of an
increasing list of examples where some previous approaches gave unsatisfactory results.
However, this informal comparison method started to be confusing
as other critical programs were found after each time a new proposal had been suggested. In the end,
it appeared that none could provide intended results for the entire list, and worse,
some disagreement on the understanding of programs occurred.
To that end, Cabalar et al.\ \cite{CabalarAEEL20} 
introduced some formal criteria, that are inherited from $\ASP$, so as to facilitate the search of a successful semantics.
Although there are newly-emerging objections \cite{Eiter.tooStrong2020}
to their soundness (even at the $\ASP$ level),
to us, that was a significant initiative to extend $\ASP$'s some well-known structural properties to the epistemic case in order 
to formally support a possible semantics proposal. We here slightly discuss $\ASP$'s possible foundational problems,
and accordingly, the validity of these properties. We mainly aim at enhancing $\ASP$'s expressivity by epistemic modalities,
and while doing so, we basically accept GL's answer sets as our underlying semantics.
However, we partly agree that especially the epistemic extensions of such properties are under debate
and had better be improved, which is the subject of another work. 
Briefly, here, we are not in search of a new semantics, compatible with the standards offered by Cabalar et al.

In this paper, we basically make a comprehensive analysis of the previous semantics approaches of $\ES$,
revealing their (dis)advantageous points. We think that this search is important to lead the way for a successful
semantics. Particularly, we propose reflexive autoepistemic $\ASP$ ($\logic{RAEASP}$) as an alternative to Cabalar
et al.'s recent approach called autoepistemic $\ASP$ ($\logic{AEASP}$). Thus, we also use Schwarz's \cite{Schwarz92}
minimal model techniques, but propose a formalism closer in spirit to the other approaches
because in $\logic{RAEASP}$, the epistemic operator $\K$ formalises knowledge, while in $\logic{AEASP}$, it
represents belief. 
We also extend the well-known propositions of here-and-there logic ($\Ht$) to the epistemic case and use them to simplify some
complex programs in order to clarify their correct meaning. We also very roughly discuss paracoherent reasoning
for epistemic logic programs, similarly to regular $\ASP$-programs \cite{paracoherent16}.

The rest of the paper is organised as follows:
Section \ref{sec:background and related work} introduces epistemic specifications ($\ES$) and its relatively
successful semantics approaches.
%
Section \ref{subsec:reflexive AEEMs} proposes a reflexive extension of autoepistemic
$\ASP$ in order to reason about a rational agent's own knowledge rather than self-belief.
Section \ref{sec:formal tools} provides some formal tools, ensuring the reasonable behaviour
of epistemic programs:
in particular, Section \ref{sec:formal properties CABALAR} recalls formal properties of $\ES$,
suggested recently.
%
Section \ref{subsec:theorems of Eht} provides epistemic extensions of some
useful equivalences of $\Ht$. 
Section \ref{subsec:comparison} gives a detailed comparison between semantics approaches discussed in the paper
by means of examples. 
%
%
Section \ref{sec:conclusion} concludes the paper with future work plan.

\section{Background and Related Work}
\label{sec:background and related work}
In this section, we introduce epistemic specifications ($\ES$) and the semantics approaches, 
proposed so far. Since Gelfond's first version, named $\ESdoksanbir$ here, was slightly and successively refined by several authors
as $\ESdoksandort$ \cite{Gelfond94}, $\ESonbir$ \cite{gelfond2011new}, $\ESondort$ \cite{kahl2014refining}, 
$\ES'_{\scriptscriptstyle{\!16}}$ \cite{KahlLS16}, and finally $\ESonsekiz$ \cite{Kahlwvconstraints18}, we begin with recalling the latest version: 
the language of $\ES$ ($\lang{\scriptscriptstyle{\ES}}$) comprises four kinds of literals;
\emph{objective literals ($l$)},
\emph{extended objective literals ($L$)},
\emph{subjective literals ($g$)}, and
\emph{extended subjective literals ($G$)} as identified below:
\vspace{-0.6cm}
\begin{center}
\small{
$$\begin{array}{|c c c c|}
\hline
\boldmath{l}~~~~~~   &
\boldmath{L}~~~~~ ~  &
\boldmath{g} &
\boldmath{G}~~~~
\\ \hline
~~ p ~~\mid ~~\classnot p ~~~~~~ & ~~~~
~~ l ~~\mid~~ \lpnot l ~~ ~~ ~~ & ~~~~
~~ \K\, l ~~\mid~~ \M\, l ~~  ~~ ~~ & ~~ ~~
~~ g ~~\mid~~ \lpnot g ~~
\\ \hline
\end{array}$$}
\end{center}
where $p$ ranges over an infinite
set $\Prop$ of atoms. $\lang{\scriptscriptstyle{\ES}}$ has 2 negations.
\emph{Strong negation}, symbolised by \singlequote{$\classnot$}, represents direct and
explicit falsity. Weaker \emph{negation as failure}
($\naf$), denoted by \singlequote{$\notlp$}, helps us partly encode incomplete information: 
$\classnot p$ implies $\notlp p$ for an atomic $p$, but not vice versa. So, if
$\notlp p$ holds, then either $\classnot p$ is the case (i.e., $p$ is false), or $p$ is assumed false since the truth of $p$ cannot
be justified due to lack of evidence. Consequently, while double $\classnot$ vanishes, $\notlp\notlp$ does not. Also note that
$\notlp p$ can be defined as a shorthand for $\bot {\leftarrow} p$, but $\classnot p$ is not a shorthand.
$\notlp p$ reads \quotes{\emph{p is false by default}}, and $\notlp \notlp p$ means \quotes{\emph{p is not false, but its truth
cannot be guaranteed}}. Different from intuitionistic modal logics, in $\ES$, the belief operator $\M$
is the dual of the knowledge operator $\K$, i.e., $\M \eqdef \notlp \K \notlp$.

A \emph{rule} is a logical statement of the form $\mathtt{head} {\leftarrow} \mathtt{body}$. 
In particular, a rule $\mathtt{r}$ of $\ES$ has the structure
$$
l_1 \lpor  ~\ldots~ \lpor  l_m \leftarrow e_{1} ~,~ \ldots ~,~ e_n
$$
in which $\mathtt{body(r)}$ viz.\ $e_{1} , \ldots , e_n$ is made up
of arbitrary (i.e., extended objective or extended subjective) literals of $\ES$, and $\mathtt{head(r)}$ viz.\
$l_1 \lpor  \ldots \lpor  l_m$ is composed of only objective literals.
Note that \onequote{$\orlp$},
\onequote{$\leftarrow$},
and \onequote{$,$} respectively represent disjunction, reversed implication and conjunction.
When $m=0$, we suppose $\mathtt{head(r)}$ to be $\bot$ and call the rule $\mathtt{r}$ a \emph{constraint} (headless rule). In particular,
when $\mathtt{body(r)}$ is composed of exclusively extended subjective literals, we call it a \emph{subjective constraint}.
When $n=0$, we suppose $\mathtt{body(r)}$ to be $\top$ and call $\mathtt{r}$ a \emph{fact} (bodiless rule).
We usually disregard $\bot$ and $\top$ in such special rules.
An \emph{(epistemic) logic program}, abbreviated as (E)LP, is a finite collection of (epistemic) rules.

\subsection{Kahl et al.'s semantics approach (\texorpdfstring{$\ESonsekiz$}{ESonsekiz}): modal reduct w.r.t. a classical S5-model }
\label{subsec:Kahl world-views}
Given a non-empty collection $\mathcal{A}$ of valuations, let $A \in \mathcal A$ be arbitrary. Then, 
satisfaction of literals is defined as follows: for an objective literal $l$, an extended objective literal $L$,
and a subjective literal $g$,
{\small
$$\begin{array}{lcllcl}
\mathcal A, A \ESmodels l  & \text{ if }&  l \in A ;
&
\mathcal A, A \ESmodels \lpnot l  & \text{ if }&  l \notin A .
\\[\smallskipamount]
\mathcal{A},A \ESmodels \K \, L & \text{ if }&  \mathcal{A}, A' \ESmodels L ~ \text{ for every } A' \in \mathcal{A};
\hspace{1.5em}
&
\mathcal{A},A \ESmodels \lpnot g & \text{ if }& \mathcal{A}, A \not\ESmodels g.
\\
\mathcal{A}, A \ESmodels \M \,L & \text{ if }&  \mathcal{A}, A' \ESmodels L  ~ \text{ for some } A' \in \mathcal A;
\end{array}$$}
\noindent
Satisfaction of an objective literal $l$ is independent of $\mathcal A$, and
satisfaction of a subjective literal $g$ is independent of $A$. So, we can safely write
$\mathcal A \ESmodels g$ or $A \ESmodels l$.
Satisfaction of an ELP $\epispec$ is defined by:
\begin{align*}
\mathcal A, A \ESmodels \epispec \text{ \ \ if \ \ } \mathcal A, A \ESmodels \mathtt{r} ~~(\text{i.e., \ }
\quotes{\mathcal A, A \ESmodels \mathtt{body(r)} \text{~ implies~ }
\mathcal A, A \ESmodels \mathtt{head(r)}})
\end{align*}
for every rule $\mathtt{r} \in \epispec$.
When $\mathcal A, A \ESmodels \epispec$ for every $A \in \mathcal A$,
we say that $\mathcal A$ is a classical S5-model of $\epispec$. In order to
decide if $\mathcal A$ is further a world-view of $\epispec$,
we first compute the (modal) reduct $\epispec^\mathcal{A} {=} \{ \mathtt{r}^\mathcal{A} {\suchthat} \mathtt{r} \in \epispec \}$
of $\epispec$ w.r.t.\ $\mathcal{A}$, where we eliminate the modal operators
$\K$ and $\M$ according to Table \ref{table:Kahl.reduct}.
{\begin{table}
\begin{center}
\small{
\caption{Kahl et al.'s original definition of reduct, and SE's implicitly inferred reduct definition. }
\label{table:Kahl.reduct}
\begin{tabular}{|l|l l|l l|}
\hline
&
\multicolumn{2}{|c|} {\textbf{Original reduct definition of $\ESonsekiz$}}   &
\multicolumn{2}{c|} {\textbf{Implicit reduct definition of $\ESonalti$}}
\\  [0.7ex]
\hline\hline
            literal $G$ 	~~
        & if  $\mathcal A \ESmodels G$  
        & if  $\mathcal A \notESmodels G$
        & if  $\mathcal A \ESmodels G$  
        & if  $\mathcal A \notESmodels G$
\\ \hline
$\K l$ 		&  \textbf{replace by} $\pmb{l}$	   & replace by $\bot$ 	 & \textbf{replace by} $\pmb{\notlp \notlp l}$
~~ & replace by $\bot$
\\ \hline
$\M l$ 			& replace by $\top$	   & replace by $\notlp \notlp l$ ~~ & replace by $\top$	 & replace by $\notlp \notlp l$
\\ \hline
$\notlp \K l$ 			& replace by $\top$		& replace by $\notlp l$ & replace by $\top$	 & replace by $\notlp l$
\\ \hline
$\notlp \M l$ 	&  replace by $\notlp l$  & replace by $\bot$ & replace by $\notlp l$		 &  replace by $\bot$
\\ \hline
\end{tabular}}
\end{center}
\end{table}}
Therefore, $\epispec^{\mathcal A}$ is a regular (nonepistemic) $\ASP$-program.
Then, we generate the set $\epset(\epispec)$ of \emph{epistemic negations}
(literals having the form of $\notlp \K \,l $ or $\M \, l$) of $\epispec$ by transforming
each extended subjective literal appearing in $\epispec$ into one of these sorts.
As an illustration,
$\epset(\Pi'){=}\set{\notlp \K p, \M q, \notlp \K s, \M t}$ for the program
$\Pi'{=}\set{t \leftarrow \K p, \M q, \notlp \K s, \notlp \M t}$.
Next, we take the elements of $\epset(\epispec)$, satisfied by $\mathcal A$ and form the set
$ \restr{\epset(\epispec)}{\mathcal A}{=}\set{G \in \epset(\epispec) {\suchthat} \mathcal A~ {\ESmodels} G}$.
Finally, $\mathcal A$ is a \emph{world-view} of $\epispec$ if\footnote{The fixed point equation $\eqfp$ is basically
to ensure stability of truth-minimisation, but, in essence, it also accommodates kind of knowledge-minimisation: e.g., given
the rule $p \lpor q$, it only holds for
$\set{\set p, \set q}$; yet, it does not hold for $\set{\set p}$ or $\set{\set q}$.}
$\mathcal A \eqfp  \anset (\epispec^{\mathcal A})$, and
\begin{align*}
&\text{there is no classical S5-model } \mathcal A' \text{ of } \epispec \text{ such that  }
\mathcal {A}' {\stackrel{\mathsf{fp}} {=}} \anset(\epispec^{\mathcal {A}'}) \text{ and } \\
&\text{(\emph{knowledge-minimisation property w.r.t.\ epistemic negation}) }
\restr{\epset(\epispec)}{\mathcal{A}'} {\supset} \restr{\epset(\epispec)}{\mathcal A}
\end{align*}
where $\anset(\epispec)$ refers to the set of all answer sets of a nonepistemic program $\epispec$.
However, knowledge-minimisation w.r.t.\ $\epset(\epispec)$ may suggest an ambiguity when $\epispec$'s classical
S5-models $\mathcal A_1$ and $\mathcal A_2$, satisfying $\mathcal A_1 \eqfp \anset (\epispec^{\mathcal A_1})$
and $\mathcal A_2 \eqfp\anset (\epispec^{\mathcal A_2})$, give rise to
$\card{\restr{\epset(\epispec)}{\mathcal{A}_1}} {\neq} \card{\restr{\epset(\epispec)}{\mathcal A_2}}$, but
$\restr{\epset(\epispec)}{\mathcal{A}_1}$ and $\restr{\epset(\epispec)}{\mathcal{A}_2}$ are not
comparable w.r.t.\ subset relation \cite{Kahlwvconstraints18}: for such $\mathcal A_1$ and $\mathcal A_2$,
it is potential to have, for instance,
$\restr{\epset(\epispec)}{\mathcal{A}_1}{=}\set{\notlp\K p, \notlp \K q}$ and
$\restr{\epset(\epispec)}{\mathcal{A}_2}{=}\set{\notlp \K s}$. So, both $\mathcal A_1$ and $\mathcal A_2$
are world-views of $\epispec$ while $\mathcal A_1$ makes more atoms unknown, compared to $\mathcal A_2$.
Another point is that we do not follow a similar truth-minimisation attitude for $\naf$ in $\ASP$, e.g.,
$\anset(p \lpor \notlp p)=\set{\set p , \emptyset}$. While we have $\emptyset \models \notlp p$ and $\set p \not\models \notlp p$
for the unique default-negated atom $\notlp p$, we do not prefer $\emptyset$ rather than $\set p$ as it minimises truth
\quotes{more} than $\set p$. Hence, to us, knowledge-minimality per $\epset(\epispec)$ had better be revised.

The main contribution of $\ESonsekiz$ over its pioneer $\ES'_{\!\scriptscriptstyle{16}}$ as a final follow-up is 
\emph{world-view constructs}: 
$\ESonsekiz$ introduces the symbol $\stackrel{\mathsf{\scriptscriptstyle{wv}\!\!\!\!}}{\leftarrow}$
which reads \quotes{it is not a world-view if}. 
This gives us a chance to transform subjective constraints
${\leftarrow} G_1, \ldots , G_n$ into ${ \stackrel{\mathsf{\scriptscriptstyle{wv}\!\!\!\!}}{\leftarrow} } G_1, \ldots , G_n$
so that they perform analogously to how constraints affect answer-sets in $\ASP$: they (at most) rule out world-views, violating them.
Note that the semantics of $\ES'_{\!\scriptscriptstyle{16}}$ has lost this property while trying to guarantee intended
results for certain other programs.

\subsection{Shen\&Eiter's approach (\texorpdfstring{$\ESonalti$}{ESonalti}): modal reduct w.r.t.\ a set of epistemic negations}
\label{subsec:SE world-views}
Another reduct-based semantics for $\ES$ has been proposed by Shen and Eiter (SE) \cite{ShenEiter16}: 
given an ELP $\epispec$, let $\mathcal A$ be its classical S5-model, and 
let $\restr{\epset(\epispec)}{\mathcal A}$ 
be the set of all its epistemic negations, satisfied 
by $\mathcal A$ (see Sect.\ \ref{subsec:Kahl world-views}).
We first transform $\epispec$ into its reduct $\epispec^{\restr{\epset(\epispec)}{\mathcal A}}$
w.r.t.\ $\restr{\epset(\epispec)}{\mathcal A}$
by replacing every $G \in \restr{\epset(\epispec)}{\mathcal A}$ with $\top$, and every
$G \in \epset(\epispec) \setminus \restr{\epset(\epispec)}{\mathcal A}$ with
$\notlp l$ if $G{=}\notlp \K l$ and with $\notlp\notlp l$ if $G{=}\M l$.
Then, $\mathcal A$ is a \emph{world-view} of $\epispec$ if
$\mathcal A \eqfp\anset(\epispec^{\restr{\epset(\epispec)}{\mathcal A}})$, and
there is no classical S5-model $\mathcal A'$ of $\epispec$  such that 
$\mathcal {A}' \eqfp \anset(\epispec^{\restr{\epset(\epispec)}{\mathcal {A'}}})$  and
$\restr{\epset(\epispec)}{\mathcal {A'}}\supset \restr{\epset(\epispec)}{\mathcal A}.$
Clearly, the reduct definitions are where $\ESonsekiz$ and $\ESonalti$ only differ.
However,
as Table \ref{table:Kahl.reduct} shows above, it is possible to arrange an equivalent version of
SE's reduct definition, and this allows us to compare the approaches
of $\ESonsekiz$ and $\ESonalti$ more easily.

Note that $\epset(\epispec)$ includes all extended subjective literals of $\epispec$
to be taken into the reduct transformation of $\ESonsekiz$, but as encoded in the form of an epistemic negation.
So, given a candidate world-view $\mathcal A$ and a subjective literal $\K l$ appearing in $\epispec$ (but
not in the scope of $\naf$),
assume that $\mathcal A \ESmodels \K l$. Note that $\epset(\epispec)$ contains $\K l$ in the form of $\notlp \K l$, and
$\notlp \K l \not\in \restr{\epset(\epispec)}{\mathcal A}$ since $\mathcal A \notESmodels \notlp\K l$. As a result,
$\notlp \K l$ is transformed into $\notlp l$ w.r.t.\ SE's reduct definition; yet the literal appears as
$\K l$ in the program $\epispec$. SE considers $\K l$ and $\notlp \notlp \K l$ to be equivalent, so since
they can transform $\notlp (\notlp \K l)$ into $\notlp (\notlp l)$, they also accept the reduct of $\K l$ into $\notlp \notlp l$
to be legitimate. Moreover, in their original definition, $\notlp \notlp l$ is reduced to $l$ in this case. To sum up, when $\mathcal A
\ESmodels \K l$, the SE-reduct transforms $\K l$ into $l$. While the other cases are reasonable, this case is not cogent for us. There are
two problematic issues here: first, the original language 
of $\ESonalti$ does not contain the modal operators as primitives, instead it has three negations; $\classnot$, $\notlp$, and $\Notlp$,
where the last denotes epistemic negation $\notlp \K$. Thus, $\K$ and $\M$ exist as derived operators
respectively in the form of $\notlp \Notlp$ and $\Notlp\notlp$. Such derivations use the equivalence between
$\K l$ and $\notlp \notlp \K l$. 
In our opinion,
$\K l$ and $\notlp \notlp \K l$ are classically equivalent, similarly to the $\ASP$-literals,
$l$ and $\notlp \notlp l$; yet, they cannot be considered strongly equivalent, allowing above transitions.
In one sense, SE's language includes $\notlp \notlp \K l$ instead of $\K l$, and there is no formal way to produce $\K l$
as a derived formula. Second, while it is questionable to
reduce $\notlp \notlp l$ to $l$ while taking the reduct of $\K l$,
replacing $\K l$ by $\notlp \notlp l$ in the reduct definition of $\ESonalti$ is probably harder to accept.
Generally speaking, taking the
reduct of a positive construct $\K l$ may be dangerous. We discuss the issue in \cite{Su19jelia} and 
propose an alternative reduct definition of $\ES$, oriented only to remove $\naf$, aligning with the approach of $\ASP$.
In particular, we do not take the reduct of $\K l$.
To sum up, although $\ESonsekiz$ and $\ESonalti$ look different, 
they are similar structurally and give the same results under SE's original reduct definition. 

The following semantics for ELPs are a lot different from the reduct-based approaches, mentioned above. They are
defined on a purely logical domain as extensions of equilibrium models\footnote{A first step towards
epistemic equilibrium logic belongs to \cite{WangZ05nested}, which embeds $\ESdoksandort$, but $\ESdoksandort$ is obselete today.}.

\subsection{Fari\~{n}as et al.'s approach (\texorpdfstring{$\ESonbes$}{ESonbes}): autoepistemic equilibrium models (AEEMs)}
\label{subsec:EEL.FHS}
Here-and-there logic ($\Ht$) 
is a 3-valued monotonic logic,
which is intermediate between classical logic and intuitionistic logic. An HT-model is an ordered pair
$(H,T)$ of valuations $H,T \subseteq \Prop$, satisfying $H \subseteq T$.
Equilibrium logic ($\EL$) is a general purpose nonmonotonic formalism,
whose semantics is based on a truth-minimality condition over HT-models.
Pearce \cite{Pearce06} basically proposed $\EL$ in order to provide a purely logical
foundation of $\ASP$. Inspired by its success as $\ASP$'s general framework, Fari\~{n}as et al.\ \cite{Su15,ijcaiFHS15,SuFH20}
introduced an epistemic extension of $\EL$, named $\ESonbes$ here, in order to suggest an alternative semantics not only for $\ES$,
but also for nested ELPs. This section briefly recalls the approach of $\ESonbes$. 

\subsubsection{Epistemic here-and-there logic (\texorpdfstring{$\Eht$}{Eht}) and its equilibrium models}
\label{subsubsec:EHT.FHS}
$\Eht$ extends $\Ht$ with nondual epistemic modalities $\K$ and $\Khat$: both operators are primitive; while $\K$
is identical to $\K {\in} \lang{\scriptscriptstyle{\ES}}$, the belief operator $\Khat$ (read \doublequotes{\emph{believed}})
is so different from $\M {\in} \lang{\scriptscriptstyle{\ES}}$. This is
justified by the fact that $\M$ is derived as $\notlp \K \notlp$ in $\ES$ and so
translated to $\Eht$ as $\intnot \K \intnot$ where $\intnot$ refers to EHT-negation. As will be shown
later in Sect.\ \ref{subsec:comparison}, $\intnot \K \intnot \phi$, $\intnot \intnot \Khat \phi$, and
$\Khat \intnot \intnot \phi$ are
all equivalent in $\Eht$. Thus, $\M {\in} \lang{\scriptscriptstyle{\ES}}$ corresponds to
$\notlp \notlp \Khat$ or $\Khat \notlp \notlp$ in an extension of $\ES$ with $\Khat$.
Notice that the difference between $\M p$ and $\Khat p$ in $\ES$ resembles
that of $\notlp \notlp p$ and $p$ in $\ASP$. As a result,  in an extended language,
we expect $\M p$ not to have a world-view,
whereas $\set{\emptyset, \set p}$ is one understandable world-view for $\Khat p$.
The language of $\Eht$ ($\lang{\scriptscriptstyle{\Eht}}$) is given by the following grammar:
for $p \in \Prop$,
\begin{align*}
\phi & \bnf p \mid \bot \mid \phi \land \phi \mid \phi \lor \phi \mid
\phi \intimp \phi \mid \K \phi \mid \Khat \phi.
\end{align*}
As usual, $\intnot \phi$, $\top$, and $\phi {\leftrightarrow} \psi$ respectively abbreviate $\phi {\intimp} \bot$,
$\bot {\intimp} \bot$, and $(\phi {\intimp} \psi) {\land} (\psi {\intimp} \phi)$.
A theory is a finite set of formulas.
An ELP $\epispec$ is translated into the corresponding
EHT-theory $\trOf{\epispec}$ via a map $\trOf{(.)}$: given a prototypical program
$\epispec {=}  \Set{r_1, r_2}$ where $ r_1=p \lpor \classnot q {\leftarrow} \M r, \notlp  s$
and $r_2= q {\leftarrow} \notlp \K p$, we have:
\begin{align*}
\trOf{\epispec} = \big( (\intnot \K \intnot r \land \intnot s ) \intimp ( p \lor \negof{q} )  \big)
~\land~ \big(\intnot \K p \intimp q \big)
~\land~ \intnot \big(q \land \negof{q} \big)
\end{align*}
where 
$\classnot q$ is evaluated as a new atom $\negof{q} \in \Prop$,
entailing the formula $\intnot \big(q \land \negof{q})$ to be inserted into \emph{}$\trOf{\epispec}$.

An EHT-model $\tuple{\there, \weak}$ is a refinement of a classical S5-model $\mathcal A$
in which valuations $A \in \mathcal A$ are replaced by HT-models $(\weak(A), A)$ w.r.t.\ a
function $\weak {\suchthat} \there \rightarrow 2^{\Prop}$,
assigning to each $A \in \there$ one of its subsets, i.e., $\weak(A)\subseteq A$. We call $\weak$
a \emph{subset} function. Thus, $\tuple{\there, \weak}$
is represented explicitly by $\big\{\big(\weak(A),A\big)\big\}_{\scriptscriptstyle{A \in \there}}$.
Satisfaction of a formula $\phi \in \lang{\scriptscriptstyle{\Eht}}$ is defined recursively
w.r.t.\ to the following truth conditions:
{\small
$$\begin{array}{lll}
 \hspace{-1em} \tuple{\there, \weak},A \Ehtmodels p   & \text{ if } & \hspace{-0.5em} p \in \weak(A);
\\
\hspace{-1em} \tuple{\there, \weak},A \Ehtmodels \phi {\intimp} \psi \hspace{-0.5em}   & \text{ if } & \hspace{-0.6em}
\big (\tuple{\there, \weak},A {\notEhtmodels} \phi \text{ or } \tuple{\there, \weak},A {\Ehtmodels} \psi \big )
\text{ and }
\big ( \tuple{\there, id},A {\notEhtmodels} \phi  \text{ or }  \tuple{\there, id},A {\Ehtmodels} \psi \big );
\\
\hspace{-1em} \tuple{\there, \weak},A \Ehtmodels \K \phi & \text{ if } & \hspace{-0.5em}
\tuple{\there, \weak},A' \Ehtmodels \phi \text{ for every } A' \in \there;
\\
\hspace{-1em} \tuple{\there, \weak},A \Ehtmodels \Khat \phi  & \text{ if } & \hspace{-0.5em}
\tuple{\there, \weak},A' \Ehtmodels \phi$ for some $A' \in \there;
\end{array}$$}
\hspace{-0.6em} where $id$ denotes the identity function.
Those of $\bot$, $\land$ and $\lor$ are standard. The EHT-model
$\tuple{\mathcal A,id}$ is called \emph{total} and identical to the classical S5-model $\mathcal A$.
Then, $\mathcal A$ is
an \emph{epistemic equilibrium model} (EEM) of  $\phi \in \lang{\scriptscriptstyle{\Eht}}$
if $\mathcal A$ is a classical S5-model $\phi$ and satisfies the following truth-minimality condition:
\begin{align} \label{minimlity of EEMs}
\text{for every possible subset function } \weak \text{ on } \mathcal A \text{ with } \weak \neq id,
\text{ there is } A \in \mathcal A
 \text{ s.t.\ }
\tuple{\there, \weak}, A \notEhtmodels \phi .
\end{align}
EEMs can only minimise truth (similarly to that of $\EL$). They do not involve a
knowledge-minimisation criterion. So, the EEM approach may bring out undesired results,
especially in the presence of disjunction. To overcome this problem,
$\ESonbes$ uses a \emph{selection process} over EEMs by comparing them  with each other according to
set inclusion $\subseteq$, and a $\phi$-indexed preorder $ \leq_\phi$ defined
as follows: for
$\there, \thereb \in \eemset(\phi)$, 
\begin{align*}
 \there \leq_\phi \thereb \text{ ~ iff ~ }
\text{for every } A_0 \in \bigcup \eemset(\phi) ,
\text{ if } \there \cup \set{A_0}, \there \starmodels \phi
\text{ then } \thereb \cup \set{A_0}, \thereb \starmodels \phi
\end{align*}
where $\eemset(\phi)$ denotes the set of all EEMs of $\phi$, and $\bigcup \eemset(\phi)$ is their union. 
Moreover\footnote{Given $\mathcal A \subseteq \mathcal B$, the pair $(\mathcal B, \mathcal A)$ denotes a multipointed S5-model
where each $A \in \mathcal A$ is a
designated (actual) world. Similarly, $(\tuple{\mathcal B, \weak}, \mathcal A)$ denotes a multipointed
EHT-model where $\tuple{\mathcal A, \weak}$ is the collection of designated HT-models of $\tuple{\mathcal B, \weak}$.},
$\there \cup \set{A_0}, \there \starmodels\! \phi$ means
$\there \cup \set{A_0}, A \sfivemodels\! \phi$ for every $A \in \mathcal A$, and
$\tuple{\there \cup \set{A_0}, \weak}, \there \notEhtmodels \! \phi$
for every $\weak \neq id$ such that $\weak(A_0)=A_0$. Then, the strict version of $\leq_{\phi}$ is standard:
$\mathcal A <_\phi \mathcal A'$ if
$\mathcal A \leq_\phi \mathcal A'$ and $\mathcal A \nleq_\phi \mathcal A'$.
An \emph{autoepistemic equilibrium model} (AEEM) of $\phi$ is the maximal EEM of $\phi$ w.r.t.\ these orderings. However,
choosing AEEMs w.r.t.\ simultaneously performing two orderings may be dangerous. So,
$\ESonbes$ should guarantee via a formal proof that these orderings do not contradict each other because it seems possible, in principle,
to have $\mathcal A_1, \mathcal A_2 \in \eemset(\phi)$, satisfying both $\mathcal A_1 \subset \mathcal A_2$ and
$\mathcal A_2 <_\phi \mathcal A_1$. Moreover, the definition of $\leq_{\phi}$ is too heavy to grasp the intuition behind.
While the preorder $\leq_{\phi}$ gets inspiration from Moore's autoepistemic logic \cite{Moore83} and Levesque's all-that-I-know logic
\cite{levesque90}, it does not use the exact techniques of these formalisms to maximise ignorance. Instead, $\ESonbes$ checks
its candidate S5-models $\mathcal A_1, \mathcal A_2 \in \eemset(\phi)$ in doubles by first enlarging them with a possible world
$A_0$ appearing in some model of $\eemset(\phi)$ and then comparing their
behaviour relative to $\phi$. Note that while testing them, if the enlarged model $(\mathcal A_1 \cup \set{A_0}, \mathcal A_1)$
is a multipointed EEM of $\phi$, then this is an advantage for $\mathcal A_1$ on the way to jump the maximality test, but
it also means that $\mathcal A_1$ is not stable w.r.t.\ knowledge in one sense. Thus,
while this tool eliminates undesired models in many cases, it does not fulfill the requirement of being understandable
in my opinion and appears a bit ad hoc. Still, $\ESonbes$ is the first formalism that has provided a \quotes{standard}
epistemic extension of $\EL$
and together with \cite{WangZ05nested}, leads the way to more successful follow-ups such as $\ESyirmi$. 
The following section introduces Cabalar et al.'s recent semantics proposal called $\ESyirmi$. 

\subsection{Cabalar et al.'s approach (\texorpdfstring{$\ESyirmi$}{ESyirmi}): founded autoepistemic equilibrium models}
\label{subsec:FAEEMs}
\emph{Autoepistemic logic} ($\ael$) \cite{Moore83} is one of the major
types of nonmonotonic reasoning, allowing a rational agent to reason about her own beliefs.
Inspired by $\ael$\footnote{Schwarz \cite{Schwarz92} showed that the nonmonotonic extensions of modal
logic $\KD$ and modal logic $\SW$ under the minimal-model semantics respectively correspond to $\ael$ and reflexive $\ael$ ($\rael$),
interpreted by stable expansions.}, $\ESonbes$ adds a valuation to EEMs
and examines the behavior of augmented models to determine AEEMs.
However, this method does not coincide with $\KD$'s minimal-model techniques because 
the AEEM-selection process takes place in an $\sfive$-setting. 
From this respect, Cabalar et al.'s approach \cite{CabalarAEEL20}, named $\ESyirmi$ here, is the first to formally combine $\EL$ and $\ael$
with the purpose of inserting the introspective
reasoning of the latter into the former. To distinguish the similar concepts of $\ESonbes$ and $\ESyirmi$, 
when necessary, we respectively add the subscripts 15 and 20.

The language $\lang{\scriptscriptstyle{\EHTyirmi}}$ is 
a fragment of $\lang{\scriptscriptstyle{\EHTonbes}}$, excluding $\Khat$, but also $\K \phi$
reads differently: \emph{$\phi$ is the agent's belief}. Semantically, it is straightforward to extend $\EHTyirmi$
with $\Khat$, but its meaning is not obvious. 

There are two important differences of $\EHTyirmimodel$-models from functional $\EHTonbesmodel$-models defined above:

First, $\EHTyirmimodel$-models are almost the same as \emph{relational} $\EHTonbesmodel$-models (see \cite{SuFH20}, Sect.\ 8)
when we consider them simply as nonempty collections of arbitrary HT-models, but disregard the relations between these HT-models.
Probably, the only (negligible) difference is that the latter can be formed as a multiset of HT-models.
In order to achieve this,
instead of a subset function $\weak$, $\EHTyirmi$ employs a serial subset relation (i.e., a multivalued subset function)
$\weakrel$, relating each $A \in \mathcal A$
to at least one element from $2^A$. So, using the S5-model $\mathcal A$ and $\weakrel$, we can produce the HT-model collections
$\set{(H,A) {\suchthat} H \weakrel A}_{A \in \mathcal A}$.
For instance, while the S5-model $\set{A}$, for $A=\set{p , q}$, can give rise to the
functional $\EHTonbesmodel$-models $\set{ (\emptyset , A)}$,  $\set{(\set p, A)}$,
$\set{(\set q , A)}$, and $\set{(A, A)}$, in $\EHTyirmi$, we can additionally obtain the
following nontotal $\EHTyirmimodel$-models
$\set{(\set p , A), (\set q, A)}$, $\set{ (\emptyset, A) , (\set p , A), (\set q, A)}$, $\set{ (\emptyset , A) , (A, A)}$, etc.
We represent $\EHTyirmimodel$-models with a similar notation $(\mathcal {A}, \weakrel)$ where $\weakrel$ refers to a
multivalued subset function on a domain $\mathcal A$.

Second, $\EHTyirmimodel$-models are in the form of $\KD$-models, while $\EHTonbesmodel$-models are special S5-models.
Given nonempty collections $\mathcal A, \mathcal B \subseteq 2^{\Prop}$ of valuations
with $\mathcal {A} \subseteq \mathcal B$ and a multivalued subset function $\weakrel$
defined on a domain $\mathcal B$,
a $\KD$-model $\tuple{\mathcal B, \weakrel}$ is a weaker form of an S5-model
$\tuple{\mathcal {A}, \restr{\weakrel}{\mathcal {A}}}$ as it may contain an additional world $(\weakrel(B),B)$
for $B \not\in \mathcal A$,
outside the maximal-cluster structure $\tuple{\mathcal {A}, \restr{\weakrel}{\mathcal {A}}}$.
Note that $\restr{\weakrel}{\set{B}}$ is an ordinary (singlevalued) subset function.
Furthermore, while $(\weakrel(B),B)$
relates exclusively to all worlds of the maximal-cluster $\tuple{\mathcal {A}, \restr{\weakrel}{\mathcal {A}}}$ and so is irreflexive,
no world in $\tuple{\mathcal {A}, \restr{\weakrel}{\mathcal {A}}}$ can relate to $(\weakrel(B),B)$.
In other words, an $\EHTyirmimodel$-model is a refinement of a classical $\KD$-model,
whose valuations are replaced by HT-models w.r.t.\ the multivalued subset function $\weakrel$.
Hence, when $\weakrel=id$, 
$\tuple{\mathcal B, \weakrel}$ corresponds to the classical $\KD$-model
$\mathcal B$. When $\mathcal A {\subset} \mathcal B$ where $\mathcal A$ is a maximal cluster,
we say that $\tuple{\mathcal B, id}$
is a \emph{proper KD45-extension} of $\tuple{\mathcal {A}, id}$.
Truth conditions of $\EHTyirmi$ only differ from those of $\EHTonbes$ for $\K \phi$ and $\Khat \phi$
at the world $(\weakrel(B),B)$: 
(in an explicit representation, we underline the world $(\weakrel(B),B)$ in the $\EHTyirmimodel$-model
$\tuple{\mathcal B, \weakrel}$
to separate it from the elements of the maximal cluster
$\tuple{\mathcal A, \restr{\weakrel}{\mathcal A}}$.)
{\small
$$\begin{array}{lll}
\tuple{\mathcal B, \weakrel}, B \Ehtmodelsyirmi \K \phi   &\text{if}&  (\mathcal {A},
\restr{\weakrel}{\mathcal {A}}), A
\Ehtmodelsyirmi \phi \text{ \ for every } A \in \mathcal {A};
\\
\tuple{\mathcal B, \weakrel}, B \Ehtmodelsyirmi \Khat \phi   &\text{if}&  (\mathcal {A},
\restr{\weakrel}{\mathcal {A}}), A
\Ehtmodelsyirmi \phi \text{ \ for some } A \in \mathcal {A}.
\end{array}$$}
\hspace{-0.6em}
Notice that since $\weakrel$ is a multivalued function on the domain $\mathcal A$, the designated world $A$ in the
above compact representation of the (pointed) $\EHTyirmimodel$-model $(\tuple{\mathcal A, \restr{\weakrel}{\mathcal {A}}}, A)$
is regarded as a shorthand for all possible HT-models $(H,A) \in \weakrel$.
The truth-minimality condition of $\ESyirmi$ is so more restricted than that of $\ESonbes$ (see \ref{minimlity of EEMs}):
for every possible multivalued subset function $\weakrel$ on the domain $\mathcal B$ satisfying $\weakrel \neq id$,
\begin{align} \label{minimlity of EEMs of ES20}
\text{ there exists } T \in \mathcal B \text{ such that } \tuple{\mathcal B, \weakrel}, T \notEhtmodels \phi
\end{align}
which amounts to saying that $\phi$ is not satisfied at the world $(H,T)$ where $H \weakrel T$ in an explicit representation
of the model $\tuple{\mathcal B, \weakrel}$. To distinguish the similar definitions, we call the condition (\ref{minimlity of EEMs of ES20})
\emph{relational} truth-minimality and the condition (\ref{minimlity of EEMs}) f\emph{unctional} truth-minimality.
Then,
an \emph{epistemic equilibrium model} ($\EEMyirmi$) of $\phi {\in} \lang{\scriptscriptstyle{\Eht}}$ is
its classical $\KD$-model 
satisfying the truth-minimality condition
(\ref{minimlity of EEMs of ES20}). Thus, when we restrict $\eemsetyirmi(\phi)$ to S5-models, $\eemsetonbes(\phi)$ is a superset of
$\eemsetyirmi(\phi)$ as the former has a more tolerant truth-minimality condition. However, in general, they are incomparable
since the latter may additionally include members in the $\KD$-model form, still remember that world-views are S5-models. Finally,
to guarantee knowledge-minimisation, $\ESyirmi$ selects S5-models in $\eemsetyirmi(\phi)$,
which has no proper $\KD$-extension in $\eemsetyirmi(\phi)$ and calls them
\emph{(founded) autoepistemic equilibrium models}\footnote{We describe the special models of the
$\ESyirmi$-semantics in a slightly different but equivalent way for ease of comparison.}
($\AEEMyirmi$) of $\phi {\in} \lang{\scriptscriptstyle{\Eht}}$.

\subsection{Our slightly new approach (\texorpdfstring{$\ESyirmibir$}{ESyirmibir}): reflexive autoepistemic equilibrium models } 
\label{subsec:reflexive AEEMs}
Modal logic $\SW$ is a reflexive closure of the modal logic $\KD$ \cite{Su17RAEL,SU.FI20}.
Schwarz proposed $\rael$ (aka, nonmonotonic $\SW$ under the minimal-model semantics) as an alternative to $\ael$ in a way that it
has $\ael$'s all attractive properties. Differently, $\rael$  defines the modality $\K$
so as to model \emph{knowledge} (which limits cyclic arguments) rather than self-belief
(which allows them) as in $\ael$. Moreover, \cite{MarekT93} discusses that $\rael$ captures the default reasoning
of $\ASP$ much better than $\ael$. Thus, $\ESyirmi$ requires a more
thorough analysis for the choice of $\KD$ rather than $\SW$ to ensure knowledge-minimisation.
This section addresses this issue and presents \emph{reflexive autoepistemic equilibrium models} (RAEEMs).

We first describe the underlying base of the new formalism $\ESyirmibir$. Similarly to $\EHTyirmi$, $\Ht$ and $\SW$
are incorporated into a monotonic formalism, referred to as $\EHTyirmibir$ hereafter. The only difference of an
$\EHTyirmibirmodel$-model from an $\EHTyirmimodel$-model is that now any HT-model $(H,T)$ in the collection is reflexive,
i.e., every such $(H,T)$ can see (access) its own information.
Relatedly, an $\EHTyirmibirmodel$-model $\tuple{\mathcal A, \weakrel}$ is formed from an $\SW$-model by
modifying its classical models (valuations) with HT-models.
When $\tuple{\mathcal A, \weakrel}$ is total, i.e., $\weakrel$ equals the identity function $id$,
we identify the $\EHTyirmibirmodel$-model $\tuple{\mathcal A, \weakrel}$ with the classical $\SW$-model $\mathcal A$.
As a result,  different from $\EHTyirmi$, $\K \phi {\intimp} \phi$ (\emph{reflexivity}) is an axiom of $\EHTyirmibir$.
The proper\footnote{Extending a cluster $\tuple{\mathcal A,\weakrel}$ to an SW5-model with an HT-model, already
existing in $\tuple{\mathcal A,\weakrel}$ does not affect satisfaction.} SW5-extension of a maximal-cluster to an
$\SW$-model is defined straightforwardly.
Given that $\mathcal B$
is a proper $\SW$-extension of a cluster $\mathcal {A}$, viz.\ $\mathcal B$ is not a cluster, truth conditions of $\EHTyirmibir$ only vary
from those of $\EHTyirmi$ for $\K \phi$ and $\Khat \phi$ at $(\weak(B),B)$ for $B \in \mathcal B \setminus \mathcal A$,
located outside the maximal cluster $\tuple{\mathcal A, \restr{\weak}{\mathcal A}}$.
{\small
$$\begin{array}{lll}
\tuple{\mathcal {B}, \weak}, B \Ehtmodelsyirmibir \K \phi   &\text{if}&  (\mathcal {B}, \weak), T
\Ehtmodelsyirmibir \phi \text{ \ for every } T \in \mathcal {B};
\\
\tuple{\mathcal {B}, \weak}, B \Ehtmodelsyirmibir \Khat \phi   &\text{if}&  (\mathcal {B}, \weak), T
\Ehtmodelsyirmibir \phi \text{ \ for some } T \in \mathcal {B}.
\end{array}$$}
\hspace{-0.6em}
The definition of (A)EEM is adjusted to the $\SW$-setting straightforwardly:
an \emph{epistemic equilibrium model} ($\eemsetyirmibir$) of $\phi \in \lang{\scriptscriptstyle{\Eht}}$ is the
classical $\SW$-model $\mathcal A$ of $\phi$, 
satisfying the truth-minimality condition
(\ref{minimlity of EEMs of ES20}), when viewed as a total $\EHTyirmibir$-model $\tuple{\mathcal A, id}$.
Similarly to $\ESyirmi$, to minimise knowledge (in other words, to maximise ignorance),
$\ESyirmibir$ also selects $\sfive$-models of $\eemsetyirmibir(\phi)$,
which has no proper $\SW$-extension in $\eemsetyirmibir(\phi)$ and calls them
\emph{reflexive autoepistemic equilibrium models} (AEEM21) of $\phi$.

\section{Some formal tools towards a well-formed epistemic extension of \texorpdfstring{$\ASP$}{ASP}}
\label{sec:formal tools}
This section first recalls the fundamental principles of $\ES$, which are still under question. Then, we demonstrate
some validities of $\EHTonbes$ that will be useful for deciding understandable models of ELPs.

\subsection{Foundational properties of \texorpdfstring{$\ES$}{ES} establishing a formal base for successful semantics}
\label{sec:formal properties CABALAR}
Since its introduction in 1991, plenty of semantics proposals have emerged for $\ES$.
However, debates and struggles to overcome unintended results still continue.
This shows that finding a satisfactory semantics of $\ES$ is a challenging task, and therefore,
as first realised by Cabalar et al.,
we need some formal support so as to reveal understandable results and wipe out undesired ones.
To this end, they proposed epistemic
splitting property (ESP), subjective constraint monotonicity (SCM), foundedness property (FP), supra-$\ASP$,
and supra-$\sfive$.
Expectedly, $\ESyirmi$ is compatible with all these properties, whereas each $\ESx$ for $x \in \set{15,16,18}$,
satisfies the last two only.
We do not reproduce the definitions here due to space restrictions, and the reader is referred to \cite{CabalarAEEL20}.
Some researchers come up with opposing arguments against their robustness
\cite{Eiter.tooStrong2020}, so a thorough examination of these tools is left to another paper. We here
check their solidity only roughly, and before doing so, we introduce these principles shortly and informally.

ESP allows for a kind of modularity that guarantees a reasonable behaviour of programs whose subjective literals are stratified.
The idea is to separate a program $\Pi$ into two disjoint subprograms (if possible), \emph{top} and \emph{bottom},
such that top queries bottom through its subjective literals, and
bottom never refers to the objective literals of top. If splitting is the case w.r.t.\ a set $U$ of literals (called \emph{splitting set}),
then we calculate the world-views of $\Pi$ in four steps: first, we
compute the world-views $\mathcal A_b$ of bottom; second,
for each $\mathcal A_b$, we take a kind of partial reduct $\Pi^{\mathcal A_b}_U$
by replacing the subjective literals $g$ (whose literals are included in $U$) of top
with their truth values in $\mathcal A_b$ (i.e., $\top$ if $\mathcal A \ESmodels g$; $\bot$ otherwise);
third, we find the world-views $\mathcal A_t$ of $\Pi^{\mathcal A_b}_U$ and end with a solution $\tuple{\mathcal A_b,\mathcal A_t}$
for $\Pi$; finally, we concatenate the components of $\tuple{\mathcal A_b,\mathcal A_t}$ in a specific way,
resulting in the world-views of the original program $\epispec$.

SCM is a special case of ESP and regulates the functioning of subjective constraints: 
when a subjective constraint $\mathtt{r}$ is added to a program $\epispec$, it
at most rules out 
the world-views of $\epispec$, but never generates new solutions, i.e., $\epispec \cup \set{\mathtt{r}}$ cannot
have a world-view $\mathcal A$, where $\mathcal A$ is not a world-view of $\epispec$ per SCM.

FP provides a derivability condition, ensuring 
self-supported world-views of a program to be rejected.

Supra-ASP means that the unique world-view of a (nonepistemic) regular $\ASP$ program $\epispec$
is the set of all its answer sets, if they exist; otherwise, $\epispec$ has no world-views.
Supra-S5 says that any world-view of an epistemic logic program is an S5-model.
Below is an example, illustrating them all.
\begin{example} [\normalfont discussed by Cabalar et al.\ \cite{CabalarAEEL20} and
Shen\&Eiter \cite{Eiter.tooStrong2020} with opposing claims] \label{ex1}
~~ \\
\normalfont
Let $\Psi=\set{\mathtt{r}_1,\mathtt{r}_2,\mathtt{r}_3}$ and $C=\set{\mathtt{r}_4}$ be the epistemic logic programs (ELPs), consisting of the rules:
\begin{align*}
\mathtt{r_1}= a \lpor b. \hspace{3em} \mathtt{r_2}= a \leftarrow \K \, b. \hspace{3em}
\mathtt{r_3}=b \leftarrow \K \, a. \hspace{3em}  \mathtt{r_4}=  \bot \leftarrow\lpnot \K \, a.
\end{align*}
As agreed by the majority, $\Psi$ has a unique world-view $\set{\set a,\set b}$ due to
knowledge-minimisation. Note that $\set{\set a, \set b}$ fails to satisfy $\mathtt{r_4}$. Thus, with SCM being applied,
$\Psi'{=}\Psi \cup C$ has no world-view. However, each $\ESx$ for $x \in \set{15,16,18}$, produces
the unique world-view/AEEM $\mathcal A{=}\set{\set{a,b}}$ for $\Psi'$.
As SCM is a special case of ESP, their result contradicts both properties. Moreover,
$\mathcal A$ also conflicts with FP since $\Set{\tuple{\set a, \mathcal A}, \tuple{\set b, \mathcal A}}$
is an unfounded set.
On the other hand, Cabalar et al.\ have already proved in separate papers that
$\ESyirmi$ satisfies all three properties above. Thus, $\ESyirmi$ follows their result and yields no AEEMs for $\trOf{(\Psi')}$.
Thanks to its relational minimality condition (\ref{minimlity of EEMs of ES20}), $\ESyirmibir$ does not produce an AEEM
for $\trOf{(\Psi')}$ either: note that the only candidate $\mathcal A$ is not truth-minimal as the weaker per (\ref{minimlity of EEMs of ES20})
S5-model 
$\set{ (\set a , A), (\set b , A) }$ also satisfies $\trOf{(\Psi')}$ where $A=\set{a,b}$, so the knowledge-minimality check is
redundant. However, if we replace (\ref{minimlity of EEMs of ES20}) with the functional minimality (\ref{minimlity of EEMs})
in $\ESyirmibir$, then $\mathcal A$ becomes truth-minimal for both $\trOf{(\Psi')}$ and $\trOf{(\Psi)}$
as none of the weaker per (\ref{minimlity of EEMs}) S5-models
$\set{ (\set a , A) }$, $\set{ (\set b , A) }$, and $\set{(\emptyset, A}$ satisfies $\trOf{(\Psi')}$ or $\trOf{(\Psi)}$. As for knowledge-minimality,
neither $\mathcal A \in \eemsetyirmibir(\trOf{(\Psi')})$ nor $\mathcal A \in \eemsetyirmibir(\trOf{(\Psi)})$
has a proper $\SW$-extension in the same
sets, so that makes $\mathcal A$ an $\aeemsetyirmibir$-model 
for $\trOf{(\Psi')}$ and $\trOf{(\Psi)}$: note that
among all possible proper $\SW$-extensions $\set{\set{a,b},\underline{\set{a}}}$,
$\set{\set{a,b}, \underline{\set{b}}}$ and $\set{\set{a,b}, \underline{\emptyset}}$ of $\mathcal A$,
none of them is in $\eemsetyirmibir(\trOf{(\Psi')})$
because they are not $\EHTyirmibirmodel$-models of $\trOf{(\Psi')}$ or $\trOf{(\Psi)}$.

At this point, we need to evaluate formally if such properties (in their original form) may
indeed be too restrictive to
reveal desired solutions. For a similar informal analysis of $\Psi'$, we refer the reader to \cite{Eiter.tooStrong2020}.
To begin with, we translate $\Psi'$ into the corresponding EHT-formula $\trOf{(\Psi')}{=}(a \lor b) \land (\K b \intimp a) \land
(\K a \intimp b) \land (\intnot \intnot \K a)$, where the last conjunct $\trOf{\mathtt{r}_4}$ is EHT-equivalent to $\K \intnot \intnot a$,
i.e., $\K(\intnot a \intimp \bot)$ by
Coroll.\ 1 in \cite{SuFH20}. So, one can interpret $\mathtt{r_4}$ in the way of applying the constraint $\bot{\leftarrow} \notlp a$
\emph{everywhere}. 
Note that world-views are S5-models in which any world is designated (actual).
Thus, replacing $\K \intnot \intnot a$ by $\intnot \intnot a$ in $\trOf{(\Psi')}$ normally should not alter the result.
If our main priority is to propose a conservative extension of $\ASP$, then $\set{\mathtt{r_1},\mathtt{r_4}}$ is
expected to derive $a$ everywhere
since it performs similarly to $\set{\mathtt{r_1}, \notlp \notlp a}$ in essence. So,
$a$ automatically appears in every world of a possible model.
Then,  $r_3$ and $r_4$ guarantee $\mathcal A$ as a world-view of $\Psi'$.
Here, the tricky point is that $\ESyirmi$'s underlying monotonic base $\EHTyirmi$ uses $\KD$-models,
and $\K \phi \intimp \phi$ (\emph{the knowledge or truth axiom}) is not a theorem of $\KD$.
Thus, replacing $\K \notlp \notlp a$ by $\notlp \notlp a$ may result in serious changes in $\ESyirmi$
and is not allowed. However, the relational truth-minimality (\ref{minimlity of EEMs of ES20}) does not allow us
to produce $\mathcal A$ even for the program $\set{\mathtt{r}_1, \mathtt{r}_2,\mathtt{r}_3, \notlp \notlp a}$ either,
while functional truth-minimality (\ref{minimlity of EEMs}) does. Then, may the condition (\ref{minimlity of EEMs of ES20}) be eliminating
understandable results? To say the least, it is questionable to have no model for $\set{\mathtt{r}_1, \mathtt{r}_2,\mathtt{r}_3, \notlp \notlp a}$.

Generally speaking, $\K$ represents belief in $\ESyirmi$, whereas it
formalises knowledge or being provable in the other semantics of $\ES$. As their major distinguishing
feature, we can \emph{believe} a statement
to be true when it is false, but it is impossible to \emph{know}/\emph{prove} a false statement. Thus, $\K p$ has no world-views
in $\ESyirmi$ as $\set{\set p, \underline{\emptyset}}$ is a proper $\KD$-extension of $\set{\set p}$. Expectedly,
$\set {\set p}$ is its unique $\AEEMyirmibir$-model. However, this result is understandable
as belief on $p$ does not imply its truth.
As a result, it may not be a good idea to compare $\ESyirmi$ with the other formalisms of $\ES$, including $\ESyirmibir$. Instead,
we can categorise it separately. As for the suitable epistemic extension of $\ES$, traces from autoepistemic
logic also exist in $\ASP$. Remember that $\notlp p$ reads: $p$ is believed not to hold under the lack of evidence to drive $p$. Moreover,
characterisation of stable models in nonmonotonic $\KD$ is well-known, and there exists translations between $\ael$ and reflexive
$\ael$, preserving the notion of expansion \cite{MarekT93}.
However, the latter reflects default reasoning better. In our opinion, $\K p$ is expected to
mean in $\ES$: $p$ is derived in \emph{all} worlds. So, interpreting $\K$ as \emph{known}
may be more appropriate to us, but it should be further discussed.

From a different perspective, we can also argue that $\mathtt{r_4}$ and $\mathtt{r_1}$ are not strong enough to generate
$\K a$, which also seems reasonable. Then, we cannot expect to have a world-view. However, we can trigger
paracoherent reasoning for $\ES$, as studied in $\ASP$ \cite{paracoherent16}
if we really need to obtain an answer for the program. In this case, the literal readings of these rules are:
$a$ is \emph{assumed} to hold everywhere in the possible model, and
also $a$ or $b$ is \emph{minimally}
the case in each world of this model. Thus, the EHT-model\footnote{When we use no subscript such as $\EHTyirmi$,
$\Eht$ is accepted to be $\EHTonbes$ by default, i.e., the combination of $\sfive$ and $\Ht$.}
$\set{\set a , (\set b, \set{a,b}) }$, in which the total HT-model $(\set a , \set a)$ is simplified into the valuation $\set a$, precisely
captures the meaning of this statement, further making $\mathtt{r_3}$ and $\mathtt{r_4}$ inapplicable as desired.
We leave the use of nontotal EHT-models as a relaxation of world-views to be discussed in future work.
\end{example}

Apart from being reliable tools for $\ES$, first, ESP is not a conservative extension of $\ASP$'s standard splitting property (SSP), i.e.,
a regular ASP-program that can be nontrivially split w.r.t.\ SSP may not be splittable w.r.t.\ ESP.
Second, FP is designed to weed out unsupported world-views
of $\ESdoksanbir$ and cannot guarantee that a founded S5-model of an ELP is also its world-view. Remember that the set of
founded classical models
of an ASP-program equals the set of its answer sets. What if $\ESdoksanbir$ does not provide a world-view for an ELP,
but this result is unintended? Moreover, FP cannot ensure the well-founded classical S5-models w.r.t.\ knowledge-minimisation. Note that
$\set{\set a}$ is a founded S5-model of $a \lpor b$ w.r.t.\ FP; yet it is unintended.
Briefly, in our opinion, these properties at least need to be strengthened before we regard them as the mandatory criteria
that a semantics of $\ES$ should comply with. 

\subsection{Some interesting validities of \texorpdfstring{$\Eht$}{Eht} that are inherited from
\texorpdfstring{$\Ht$}{Ht}}
\label{subsec:theorems of Eht}
Now, we extend some well-known propositions of $\Ht$ to $\Eht$, which we use later
for a correct understanding of the real behaviour of complex programs.
First, recall that a formula $\phi \in \lang{\scriptscriptstyle{\Eht}}$ is \emph{satisfiable} 
if it has an EHT-model. If every
EHT-model satisfies $\phi$, then it is \emph{valid} (\singlequote{$\Ehtmodels \phi$}). Given 
$\phi, \psi \in \lang{\scriptscriptstyle{\Eht}}$, $\psi$ is a \emph{logical consequence} of $\phi$ in $\Eht$
(\singlequote{$\phi \Ehtmodels \psi$}) if every EHT-model of $\phi$ satisfies $\psi$. When
$\phi \Ehtmodels \psi$ and $\psi \Ehtmodels \phi$ (i.e., they have the same EHT-models),
we call them \emph{logically equivalent} in $\Eht$.
%
%
\noindent
\begin{prop} [\normalfont de Morgan laws and the weak law of the excluded middle %
both hold in $\Eht$.] \label{prop:de Morgan laws} ~~
\begin{center}
\begin{tabular}{ll}
 $\Ehtmodels \intnot (\phi \land \psi) \leftrightarrow \intnot \phi \lor \intnot \psi $  \hspace{5em}& \hspace{5em}
 $\Ehtmodels \intnot \phi \lor \intnot \intnot \phi$
 \\
 $\Ehtmodels \intnot (\phi \lor \psi) \leftrightarrow \intnot \phi \land \intnot \psi $ \hspace{5em}& \hspace{5em}
$\Ehtmodels \intnot \intnot \intnot \phi \leftrightarrow \intnot \phi$
\end{tabular}
\end{center}
\end{prop}
\begin{prop} \label{prop:equivalence} \normalfont For $\phi,\chi,\psi \in \lang{\scriptscriptstyle{\Eht}}$,
the following formulas are logically equivalent in $\Eht$: ~~
\begin{center}
$\text{i.)}  \Ehtmodels (\intnot \intnot \phi \land \chi \intimp \psi) \leftrightarrow (\chi \intimp \intnot \phi \lor \psi) \hspace{2em}
\text{ii.)}   \Ehtmodels (\intnot \phi \land \chi \intimp \psi) \leftrightarrow (\chi \intimp \intnot \intnot \phi \lor \psi)$
\end{center}
\end{prop}
%
%
\begin{coroll} \label{coroll:equivalence}
\normalfont As an immediate consequence of Prop.\ \ref{prop:equivalence} (hint: take $\chi = \top$), we have:
for $\phi,\psi \in \lang{\scriptscriptstyle{\Eht}}$,
\begin{center}
$\text{i.)} \Ehtmodels (\intnot \intnot \phi \intimp \psi) \leftrightarrow (\intnot \phi \lor \psi) \hspace{5.6em}
\text{ii.)} \Ehtmodels (\intnot \phi \intimp \psi) \leftrightarrow (\psi \lor \intnot \intnot \phi)$
\end{center}
\end{coroll}
%

\subsection{Comparison between semantics proposals of \texorpdfstring{$\ES$}{ES} via some critical examples}
\label{subsec:comparison}
As mentioned above,
AEEMs are in the form of classical S5-models. $\ESonbes$ chooses
the AEEMs of a formula $\phi$ from the set $\eemsetonbes(\phi)$ of the candidates. Differently from $\ESyirmi$
and $\ESyirmibir$,
$\ESonbes$ executes a pairwise comparison to the members 
of this set to guarantee knowledge minimisation: for instance,
when $\eemsetonbes(\phi){=}\set{\mathcal {A}_1, \mathcal {A}_2}$,  we
eliminate $\mathcal A_1$ if $\mathcal A_1 {\subset} \mathcal A_2$ or
$\mathcal A_1 {<_\phi} \mathcal A_2$, and so we get $\mathcal A_2 \in \aeemsetonbes(\phi)$. 
This strategy fails when we add a constraint which $\mathcal A_2$ violates because then
$\mathcal A_1 \in \aeemsetonbes(\phi)$ rather than having no AEEMs. On the other hand,
$\ESyirmi$ tests the members of $\eemsetyirmi(\phi)$ according to 
whether they have a proper $\KD$-extension in $\eemsetyirmi(\phi)$, and so adding constraints do not cause inconsistencies.
More explicitly, $\aeemsetyirmi(\phi){=}\set{\mathcal {A}_2}$
when $\eemsetyirmi(\phi){=}\set{\mathcal {A}_1, \mathcal {A}_2, \mathcal A_3}$ where $\mathcal A_3$
is a proper $\KD$-extension of $\mathcal A_1$. However, adding a subjective constraint which
is not satisfied by $\mathcal A_2$ causes the lack of AEEMs for $\phi$. The case for arbitrary constraints should further be checked.
Note that $\ESonalti$ and $\ESonsekiz$ also suffer from a similar pairwise comparison of possible candidates. The
following example illustrates this discussion.
\begin{example} [\normalfont given by Cabalar et al.\ \cite{CabalarAEEL20} 
to show that $\ESonbes$, $\ESonalti$, and $\ESonsekiz$ violate epistemic splitting] \label{ex2} ~~
\\
\normalfont
Let $\Sigma=\set{\mathtt{r_1}, \mathtt{r_2}, \mathtt{r_3}}$ be the epistemic logic program (ELP), consisting of the rules given below:
\begin{align*}
\mathtt{r_1}= a \lpor b. \hspace{40pt}  \mathtt{r_2}= c \leftarrow \K \,a. \hspace{40pt}
\mathtt{r_3}=\bot \leftarrow \notlp c. \hspace{10pt} ( \text{ or, } \mathtt{r'_3}= \notlp \notlp c.)
\end{align*}
First take $\Sigma_1{=}\set{\mathtt{r}_1,\mathtt{r}_2}$: it has a unique and clearly understandable world-view
$\mathcal A_1{=}\set{\set a, \set b}$ in $\ESonalti$ and $\ESonsekiz$. Note that $\mathcal A_2{=}\set{\set{a,c}}$
does not occur as a truth-minimal model of $\Sigma_1$ in $\ESonalti$ and $\ESonsekiz$, thanks to their
fixed point equations $\eqfp$.
However, in $\ESonbes$ and $\ESyirmi$, we have both $\mathcal A_1$ and $\mathcal A_2$ as truth-minimal EEMs
respectively according to the tools
\ref{minimlity of EEMs} and \ref{minimlity of EEMs of ES20}.
Fortunately, they eliminate $\mathcal A_2$
w.r.t.\ their knowledge-minimisation properties. Then, consider the whole program $\Sigma$: now,
$\ESonbes$, $\ESonalti$ and $\ESonsekiz$ all withdraw $\mathcal A_1$ since it violates the constraint $\mathtt{r_3}$ and instead
choose $\mathcal A_2$ as the unique world-view/AEEM: for $\Sigma$ and $\mathcal A_2$,
the fixed point equations of $\ESonalti$ and $\ESonsekiz$ hold, and now there is no rival.
To us, this result provided by $\mathcal A_2$ is unsupported: 
while an agent disjunctively has two alternative information, $a$ and $b$, about a world,
she cannot justify $\K a$. So, $\mathtt{r_2}$ becomes inapplicable and the existence of $c$ is unfounded.
Further inserting the constraint $\mathtt{r_3}$ can guarantee neither $\K a$ nor $c$.
Thus, $\Sigma$ should have no world-views/AEEMs as is the case in $\ESyirmi$ because
$\set{\set{a,c}} \in \eemsetyirmi(\Sigma)$ has the proper $\KD$-extension $\set{\set{a,c}, \underline{\set{b,c}}}$
in $\eemsetyirmi(\Sigma)$. As expected, Cabalar et al.'s principle of ESP aligns with the result of $\ESyirmi$. 
As $\aeemsetyirmibir(\Sigma)=\set{\mathcal A_2}$, we show by this counterexample that ESP does not hold for $\ESyirmibir$.
Note that $\mathcal A_2 \in \eemsetyirmibir(\Sigma)$ has no proper $\SW$-extension in the same set: the only candidate
does not hold as $\set{\set{a,c}, \underline{(\set b, \set{b,c})}}$ satisfies $\Sigma$.
\end{example}

\begin{example} [\normalfont used by Kahl as a motivating example for his new
modal reduct first given in \cite{kahl2014refining}] \label{ex3} ~~
\\
\normalfont
Take the ELP $\Delta{=}\set{\mathtt{r_1}, \mathtt{r_2}}$ where $\mathtt{r_1}{=} a \lpor b$
and $\mathtt{r}_2{=}b \leftarrow \M\, a$,
and then translate it into the corresponding EHT-formula
$\trOf{\Delta}{=}(a \lor b) \land (\intnot\K \intnot a \intimp b )$.
We know that $\intnot \intnot \Khat$, $\Khat \intnot \intnot$ and
$\intnot \K \intnot$ are all equivalent in $\Eht$ (see Prop.\ 5; \cite{SuFH20}) 
So, using Coroll.\ \ref{coroll:equivalence}, we deduce that $\trOf{\Delta}$
is equivalent to $(a \lor b) \land (b \lor \intnot \Khat a)$ in $\Eht$, and again by Prop.\ 5 \cite{SuFH20},
even further to $b \lor (a \land \K \intnot a)$. Note that
the last disjunct yields a contradiction in $\EHTonbes$ and $\EHTyirmibir$, making
%
$\trOf{\Delta}$ and $b$ EHT-equivalent. 
Thus, $\Delta$ has the unique AEEM $\set{\set b}$ in $\ESonbes$ and $\ESyirmibir$; yet
$\ESyirmi$ gives no AEEMs as $\set{\set b}$ has a proper
$\KD$-extension $\set{\set b, \underline{\set a}}$ in $\eemsetyirmi(\Delta)$.
%
$\Delta$ cannot be split w.r.t.\ ESP. However, $\set{\set b}$ is a founded model of $\Delta$ w.r.t.\ FP.
So, a semantics satisfying FP is supposed to have this world-view. Semantics like $\ESyirmi$ and
$\ESdoksanbir$ jump over this test
since they do not have world-views for $\Delta$. This is why we find it essential to reinforce FP 
so as to guarantee that a successful semantics should be able to bring out all founded S5-models of an ELP as its
world-views/AEEMs.
\end{example}
\begin{example} [\normalfont discussed by Cabalar et al.\ \cite{CabalarAEEL20} to show that $\ESonbes$, $\ESonalti$, and $\ESonsekiz$
violate epistemic splitting] \label{ex4} ~~
\normalfont
Let $\Upsilon=\set{\mathtt{r_1}, \mathtt{r_2}, \mathtt{r_3}, \mathtt{r_4}}$ be the epistemic logic
program (ELP), composed of the following rules:
\begin{align*}
\mathtt{r_1}= a \lpor b.  \hspace{30pt}  \mathtt{r_2}= c \lpor d \leftarrow \notlp \K \, a. \hspace{30pt}
\mathtt{r_3}=\bot \leftarrow c. \hspace{30pt}  \mathtt{r_4}=  \bot \leftarrow d.
\end{align*}
Then, $\trOf{\Upsilon}=(a \lor b) \land (\intnot \K a \intimp c\lor d)\land (\intnot c)\land(\intnot d)$.
By Prop.\ \ref{prop:de Morgan laws} and Coroll.\ \ref{coroll:equivalence},
$\trOf{\Upsilon}$ is equivalent to $(a \lor b) \land ((c \lor d) \lor \intnot \intnot \K a)
\land \intnot(c\lor d)$. Using Coroll.\ 1 in \cite{SuFH20}, 
we further simplify $\trOf{\Upsilon}$ into $(a \lor b) \land (\K \intnot \intnot a) \land \intnot (c \lor d)$.
Thus, this formula, in essence, has the same meaning
as $(a \lor b) \land (\intnot \intnot a)\land \intnot c \land \intnot d$,
whose unique world-view/AEEM is $\set{\set a}$ in each $\ESx$ for $x \in \set{15,16,18,20,21}$ w.r.t.\ supra-$\ASP$.
So, for a semantics of $\ES$ with classical S5-models (i.e., according to supra-$\sfive$), $\set{\set a}$ is expected to be the only
world-view/AEEM for $\Upsilon$. Nonetheless, $\ESyirmi$ has no AEEMs for $\Upsilon$ because the unique possibility $\set{\set a}$
has the proper $\KD$-extension $\set{\set a, \underline{\set b}} \in \eemsetyirmi(\Upsilon)$. Of course, this result is normal
because reflexivity is not valid in $\EHTyirmi$, and so it is not legal to make such transitions in it. However, we can assert that
the knowledge-minimisation technique of $\ael$ may not be the best choice to be employed in $\ES$.
Note that $\ESyirmibir$, using the reasoning of $\rael$, obtains the AEEM $\set{\set a}$ for $\trOf{\Upsilon}$
as $\set{\set a, \underline{\set b}} \notEhtmodelsyirmibir \trOf{\Upsilon}$. As an advantage, extending $\ESyirmibir$ with
world-view constructs \cite{Kahlwvconstraints18} will then make $\ESyirmibir$ more expressive than $\ESyirmi$. Also, SCM is
useful in problem descriptions of some domains like conformant planning \cite{Kahlwvconstraints18,CabalarAEEL20}.
%
\end{example}
%

\section{Conclusion}
\label{sec:conclusion}
The main purpose of this paper is to carefully revise the competing approaches of $\ES$, among which
are $\ESx$ for $x \in \set{15,16,18,20}$. We systematically bring to light the (dis)advantages 
of these formalisms. In doing so, we discuss how we can reach a more suitable epistemic extension of $\ASP$. 
We also propose a slightly new formalism called $\ESyirmibir$, which can also be regarded as reflexive $\ESyirmi$. We do so
because $\ESyirmi$ uses a well-studied technique of knowledge minimisation, but it is a nonmonotonic epistemic logic of
belief, while all the rest can be considered as epistemic formalisms of knowledge.
%
As future work, we will first establish a strong equivalence characterisation
of ELPs under the $\ESyirmibir$-semantics, which is identified as $\EHTyirmibir$-equivalence. 
Then,
we also would like to study paracoherent semantics of ELPs.




\nocite{*}
\bibliographystyle{eptcs}
\bibliography{ezgi}
\end{document}